\documentclass[11pt]{article}
\usepackage{graphicx}

\usepackage[]{EMNLP2023}

\usepackage{times}
\usepackage{latexsym}
\usepackage{makecell}
\usepackage[T1]{fontenc}

\usepackage[utf8]{inputenc}

\usepackage{microtype}
\usepackage{graphicx}
\usepackage{amsmath}
\usepackage{amsthm}
\usepackage{booktabs}
\usepackage{algorithm}
\usepackage{algorithmic}
\usepackage{mathtools}
\usepackage{booktabs}
\usepackage{multirow}
\usepackage{microtype}
\usepackage{wasysym}
\usepackage{amssymb}
\usepackage{xspace}

\usepackage{inconsolata}

\title{Social Media Ready Caption Generation for Brands}

\author{Himanshu Maheshwari, Koustava Goswami, Apoorv Saxena \and Balaji Vasan Srinivasan \\ \texttt{\{himahesh, koustavag, apoorvs, balsrini\}@adobe.com} \\ Adobe Research, India} 

\begin{document}
\maketitle
\begin{abstract}
 Social media advertisements are key for brand marketing, aiming to attract consumers with captivating captions and pictures or logos. 
 While previous research has focused on generating captions for general images, incorporating brand personalities into social media captioning remains unexplored. Brand personalities are shown to be affecting consumers' behaviours and social interactions and thus are proven to be a key aspect of marketing strategies. 
 Current open-source multimodal LLMs are not directly suited for this task. Hence, we propose a pipeline solution to assist brands in creating engaging social media captions that align with the image and the brand personalities. 
 Our architecture is based on two parts: \textbf{a} the first part contains an image captioning model that takes in an image that the brand wants to post online and gives a plain English caption; \textbf{b} the second part takes in the generated caption along with the target brand personality and outputs a catchy personality-aligned social media caption. Along with brand personality, our system also gives users the flexibility to provide \textit{hashtags}, \textit{Instagram handles}, \textit{URLs}, and \textit{named entities} they want the caption to contain, making the captions more semantically related to the social media handles. Comparative evaluations against various baselines demonstrate the effectiveness of our approach, both qualitatively and quantitatively.
\end{abstract}

\section{Introduction}
With the constant increase in consumer interest towards digital marketing, organizations are trying to position their brands uniquely in the competitive market. Brands are visioned as the media to build customer's perception about the company and their products \cite{pamuksuz2021brand}, proven to be playing a significant role to enhance the company's performance \cite{madden2006brands}. With the increase in customer engagement in social media sites, companies are trying to advertise their brands in different social media including \textit{Instagram}, \textit{Facebook} etc., to attract consumers from diverse age groups. It includes publishing their brand images with catchy captions combined with different 
\textit{hashtags} and \textit{emojis}. With the advancement of state-of-the-art deep learning models, different methodologies are used to generate captions of these images to be posted on social media. Vision-language models \cite{DBLP:conf/icml/RadfordKHRGASAM21,DBLP:conf/nips/LiSGJXH21,DBLP:conf/icml/WangYMLBLMZZY22,DBLP:conf/nips/AlayracDLMBHLMM22} pre-trained on massive image-text pairs are used to generate linguistic descriptions from image for various downstream tasks. Recently, \citet{DBLP:journals/corr/abs-2301-12597} proposed \textit{BLIP-2}, capable of state-of-the-art caption generation from images in zero-shot setup. But how to align these captions with brand features and specificity for social media posts is still hard to be achieved.  

While brand images are portrayed to be the median of attraction for consumers, attachment to brand personalities plays a key role in associating human attributes with brands \cite{xu2016predicting}. Brand personality refers to the set of human characteristics associated with a brand \cite   {doi:10.1177/002224379703400304}. In a social setting brand personalities can be associated with the consumer's social needs and self-expression \cite{markus1986possible}. Humans express their different choices in their close larger group by showing their affection to the particular brands. \citet{batra2012brand} pointed out that consumers' natural tendency is to choose brands that depict the personality aligning with their personalities. Thus to be successful in digital marketing, aligning the brand personality with the social media posts and captions, has become very essential. Following \citet{doi:10.1177/002224379703400304}, we focus on five brand personalities: Sincerity, Excitement, Competence, Sophistication, and Ruggedness. Table \ref{traits_table} shows the different traits of these brand personalities \cite{doi:10.1177/002224379703400304}.    

\renewcommand{\arraystretch}{1.2}
\setlength{\tabcolsep}{4pt}
\begin{table}
\centering \scriptsize
\begin{tabular}{ll}
\hline
\textbf{Personality} & \textbf{Traits}\\
\hline
Sincerity & Down-to-earth, Honest, Wholesome, Cheerful \\
Excitement & Daring, Spirited, Imaginative, Up-to-date\\
Competence & Reliable, Intelligent, Successful \\
Sophistication & Upper Class, Charming\\
Ruggedness & Outdoorsy, Tough\\
\hline
\end{tabular}
\caption{Personality traits}
\label{traits_table}
\end{table}

The existing research works \cite{xu2016predicting,pamuksuz2021brand} have proposed a novel methodology to predict brand personalities from Social media texts but have not inspected how to align the brand personalities to the published captions in social media. \citet{roy2021integrated} in their work investigated how to improve the brand content present on the web by integrating the brand personalities using different linguistic features but lacking in aligning to generated captions. On the other hand, how to generate consistent and catchy captions from the image of a brand is still very much unfolded. Regular single-line image captions can be generated using state-of-the-art vision-language models like \textit{InstructBLIP} and \textit{Flamingo}. However, these models fail to generate social media captions with brand personality alignment, as we will show quantitatively and qualitatively. Also, during this caption generation, there is no scope for the companies to inject their own \textit{hashtags}, \textit{social media handles}, \textit{URLs}, \textit{named entities} affecting their visibility in the social media sites. To address the above challenges, we propose a pipeline approach. 

\begin{figure*}[t!]
    \centering
    \includegraphics[width=\textwidth]{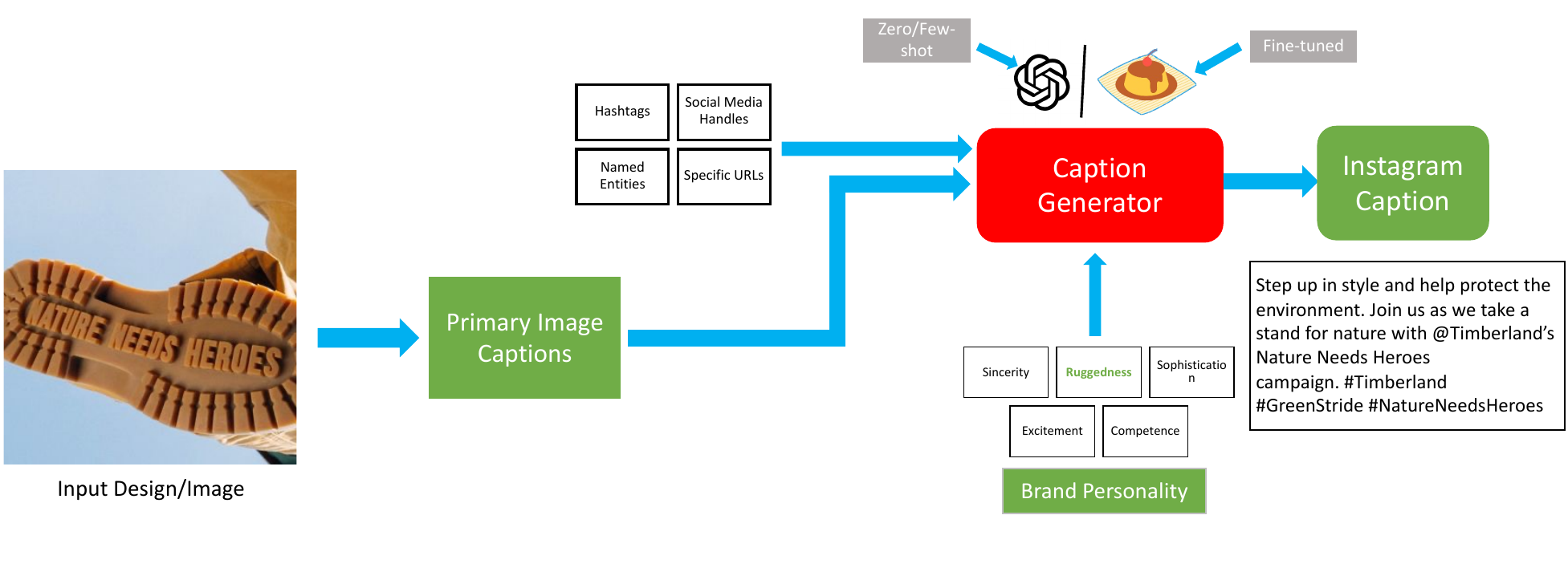}
    \caption{Architectural Diagram: It begins with an image input and generates an initial caption using BLIP-2. The caption generator module takes primary captions, personalities, and possibly other user-specific elements. This module supports GPT* for zero/few-shot scenarios and any open-source LLM for fine-tuned setups.}
    \label{fig:architecture}
\end{figure*}

Thus, we propose a new task of \textbf{generating automatic brand captions from the brand images for social media posts} while aligning with the \textbf{brand-specific personalities}. We propose a framework capable of generating catchy social media captions from the brand images aligned to the brand personality and brand specificity. Our proposed framework (refer to Figure \ref{fig:architecture}) consists of two parts: \textbf{(i)} First, we generate normal image captions in a zero-shot setup from a vision-language model from the brand images. This gives us the flexibility to plug and play any state-of-the-art vision-language model while also reduce down the need for huge computation and data resources during initial caption generation; \textbf{(ii)} second, the initially generated captions are then fed into our large language model (LLM) based brand personality alignment framework to generate the final catchy captions. During the final generation our systems optionally take  \textit{hashtags}, \textit{social media handles}, \textit{URLs}, and \textit{named entities} to make the captions more semantically related to the parent company's personality and profiling. We are releasing two versions of the framework, \textbf{(a)} \textit{fine-tuned LLM based framework}, where open sourced LLM is finetuned on the brand aligned training dataset. This provides users more flexibility in choosing frameworks according to their needs. \textbf{(b)} \textit{few-shot LLM-based framework}, where the brand alignment component consists of \textit{OpenAI's GPT}\footnote{\url{https://platform.openai.com/docs/models}} series models;  The few-shot brand alignment model makes the framework more compute efficient but costly, whereas the fine-tuned LLM-based framework, once trained, is cheap and does not require brands to share their private data with OpenAI. Finally, our framework also allows the user to generate Instagram captions even without images. The user can directly use the second part of our framework to generate social media captions.

Due to the lack of a good-quality open-source dataset, we create our dataset from public Instagram accounts. We quantitatively study the limitation of this dataset. We propose to use the reference-free G-Eval metric to measure brand alignment. Additionally, we propose a fine-tuned CLIPScore to measure the relevance of captions to the associated images.
Extensive experiments conducted on Instagram posts (Section 6) demonstrate the effectiveness of our framework. For instance, our proposed approach generates captions that are aligned with both the target personality and image. Using BLIP2 and zero-shot GPT, our pipeline achieves a clipscore of $0.887$ and $52.000\%$ of captions in the target personality. We also train a FlanT5 model that produces captions with a clipscore of $0.830$ and $42.339\%$ of captions in the target personality. Our pipeline outperforms end-to-end image-to-text approaches. The recent InstructBLIP model, utilizing one shots, achieves a clipscore of $0.878$ and an accuracy of $39.600\%\%$. Our framework allows users to provide additional attributes like hashtags, entities, usernames, and URLs, unlike InstructBLIP.

In a nutshell;
\begin{itemize}
    \item We propose a novel task of generating automatic brand captions from brand images for social media posts aligned to brand personalities and investigate how the current vision-language model fails for this task.
    \item We propose a framework capable of generating this automatic captions supporting both zero/few-shot and fine-tuning capabilities of the large language models
    \item The proposed framework also allow users to specify their desired hashtags and social media handles making the captions more aligned with the specific brand features and specificity.
    \item We identify and address the limitations of existing datasets and evaluation metrics. We do an extensive evaluation of our proposed framework.
\end{itemize}

\section{Related Works}

As we are proposing a new task of caption generation including baseline results and methodology from brand images, no such related works are found on the same pipeline. Instead here we have tried to summarise the related works involved for each section of caption generation from images and brand identification from texts.

\noindent{\textbf{Identification of Brand Personality}} 
Brand personality identification process has been explored by different researchers at different times. Earlier using different conventional empirical methods including user surveys wide range of data was collected for the brand personality testing \cite{aaker1997dimensions,carr1996cult}. As the usage of different social media sites scaled up, using different data analytic methodologies the customer insights are unveiled at different times \cite{DBLP:journals/misq/ZhangBR16}. Having said that, these customer analytical insights enhanced the scope of the research on brand personality identification and understanding \cite{DBLP:conf/socialcom/GolbeckRET11}. 
\citet{DBLP:journals/mktsci/CulottaC16} released an automated data analysis tool capable of predicting brand perceptions from tweets. \citet{ranfagni2016measure} used a parts-of-speech tagger to detect personalities from a text. \citet{DBLP:journals/mktsci/HuangL16} implemented a support-vector machine approach to detect consumer preferences related to brand personalities. \citet{PAMUKSUZ202155} trained a RoBERTa-based classifier to detect brand personality.

\noindent{\textbf{Image-to-Text Caption Generation}} Image-to-text caption generation is explored by different researchers using various deep learning and probabilistic methods. Some of the very first works generated sentences using RNN by considering sentence generation as a sequence-to-sequence problem \cite{DBLP:journals/corr/KleinLSW14,DBLP:journals/corr/MaoXYWY14a}. But this does not correlate the image patches to sentences. Thus the introduction of attention based localization was the first step to identify relevant salient objects \cite{DBLP:conf/nips/SchwartzSH17,DBLP:conf/cvpr/SchwartzYHS19}. With the introduction of the transformers model, using self-attention blocks this image localization-based attention improved over time \cite{DBLP:conf/iclr/DosovitskiyB0WZ21,DBLP:conf/cvpr/PanYLM20,DBLP:conf/cvpr/SchwartzSH19}. 
In recent times, to train the image-text model in an unsupervised way, a large scale vision-language scrapped dataset is used. This datasets are used in the pre-training stage \cite{DBLP:conf/eccv/Li0LZHZWH0WCG20,DBLP:conf/cvpr/ZhangLHY0WCG21} 
\citet{DBLP:conf/nips/AlayracDLMBHLMM22} proposed Flamingo trained by injecting new cross-attention layers into the LLM to learn visual features and pre-trained on a large set of image-text pairs. Recently, \citet{DBLP:journals/corr/abs-2301-12597} proposed BLIP-2 trained on frozen language model and image encoders for various image-text tasks including image captioning. This model is easy to load with lower computational cost and generates state-of-the-art captions from images. In our case for caption generation, we have used BLIP-2 in zero-shot settings.
\citet{attend2u:2017:CVPR} proposed a personalised image captioning task which is probably the closed work to ours. 

\section{Methodology}
This work proposes a pipeline for generating brand-specific Instagram captions. The pipeline takes an image, brand personality, and optional attributes (hashtags, URLs, named entities, usernames) as input. The output is a relevant Instagram caption that aligns with the brand personality and includes the provided attributes. Our framework consists of two parts: a) The first part generates captions from a vision-language model in a zero-shot setup, using only the brand image as input. b) The second part generates the final Instagram caption using a fine-tuned or zero-shot language model. The input to this part includes the caption from the vision-language model, brand personality, and any additional attributes provided by the user. Figure \ref{fig:architecture} shows the overall pipeline, and subsequent section provides a detailed description of each component.

\subsection{Part 1: Automatic Image Captioning}
We employ a zero-shot vision-language model to generate image captions, offering the advantage of flexibility to use any state-of-the-art model. Due to its competitive performance, we use BLIP2 (\textit{pretrain\_flant5xxl}) \cite{li2023blip2} as the vision-language model to generate image captions in our work. The BLIP2 caption achieves a remarkably high CLIPScore of 0.9255, as discussed in Section 5.1.

This pipeline approach offers flexibility to the users. In cases where the user has an initial idea but lacks an image, they can provide a one-line description instead. The subsequent part of the pipeline accepts either an image caption or the user's input. This allows users to generate Instagram captions even in the absence of corresponding images.

\subsection{Part 2: LLMs for Final Instagram Caption}
The second part of the pipeline incorporates the image caption/description, brand personality, and optional attributes (hashtags, URLs, named entities, usernames). We explore two frameworks: \textbf{(a)} a Fine-tuned LLM and \textbf{(b)} Zero/Few-shot GPT.

\textbf{Finetuned LLM: }We instruction-finetune a FlanT5-XL \cite{https://doi.org/10.48550/arxiv.2210.11416} model to generate Instagram captions. The model takes as input an instruction, including an image caption, brand personality, and additional attributes, and produces an Instagram caption as output. We train the model by optimizing cross-entropy loss between the generated caption and the ground truth caption. Since FlanT5 cannot handle emojis, we utilize the emoji python package \footnote{\url{https://pypi.org/project/emoji/}} to convert emojis into plain text, and during inference, we use the same to convert the textual representation of emoji to actual emoji. Table \ref{flant5_example} shows different instructions that we tried. 

We provide additional attributes to the user in the instruction. We explore two instruction variants, namely \textit{selective} and \textit{non-selective}, which differ in how additional attributes are provided.
The main difference between selective and non-selective is how much information we provide to the model about attributes not present in the caption. 
In the \textit{selective} variant, if a user doesn't provide an attribute, it is not mentioned in the instruction. In contrast, if the user has not provided any attribute in non-selective, then the instruction has that attribute, and its value is \textit{None}. The intuition behind this is to check how the model will perform in different levels of information being present. In selective, sometimes some attributes are mentioned in the caption, and sometimes not, whereas in non-selective, such attributes are always mentioned in the caption. Due to space restrictions, we do not show the exact instruction in the main text. Please refer to Table \ref{flant5_example} in the Appendix for exact instructions.

A fine-tuned language model offers greater output control, cost-effectiveness, improved results compared to GPT, and ensures privacy of user data.

\vspace{2.5pt}
\textbf{Zero/Few-shot GPT: }We also use GPT (\textit{gpt-3.5-turbo}) to create the Instagram caption. We experiment with zero-shot and a few shot settings. The GPT prompt includes instructions, image caption, personality, and additional attributes. In the few-shot setting, we provide a few carefully selected examples (up to four) from the training set. Initial investigations revealed that in the zero-shot setting, over 25\% of the captions contained the personality word, indicating overfitting.  For example, if the target personality is "Sophisticated," the output will have words like sophisticate, sophisticated, sophistication, etc. So in the zero-shot setting, we explicitly ask the GPT that the output should not contain the tonality word or any variant of it.

Like FlanT5 finetuning, we also experiment with two variants of GPT prompting: selective and not selective. 
Due to space restrictions, we do not show the exact prompt in the main text. Please refer to Table \ref{gpt_example} in the Appendix for one-shot prompting to GPT. GPT allows the user to create captions even if the training data is not there. 

\renewcommand{\arraystretch}{1.2}
\setlength{\tabcolsep}{4pt}
\begin{table}
\centering \scriptsize
\begin{tabular}{lccc}
\hline
\textbf{Personality} & \textbf{Train} & \textbf{Validation} & \textbf{Test}\\
\hline
Sincerity & 4,433 & 547 & 50 \\
Excitement & 5,055 & 588 & 50 \\
Competence & 2,410 & 333 & 50 \\
Sophistication & 7,389 & 867 & 50 \\
Ruggedness & 6,119 & 748 & 50 \\
\hline
\end{tabular}
\caption{Dataset Statistics}
\label{dataset_table}
\end{table}

\section{Dataset}
To address the absence of an existing dataset for our task, we scrape images and captions from public Instagram accounts. \citet{Xu_Liu_Gou_Akkiraju_Mahmud_Sinha_Hu_Qiao_2021} has conducted extensive human studies to identify the personality of 219 brands and classified them into five personality types: sincerity, excitement, competence, sophistication, and ruggedness. Using their work, we select six brands from each personality type. Only those brands are selected that occur exclusively in one personality type. Exact list of brands in each personality is present in appendix. We collect (maximum) the top 2000 posts for each brand. The cut-off date for data collection is 27 April 2023. Here we assume that all the captions in each brand will reflect the brand personality. A more critical study of this assumption and overall dataset quality is done in section 5.2. Out of these six brands, we select one brand exclusively for testing for each personality to prevent the models from overfitting brand-specific nuances. We remove captions that are just emojis, smaller than ten words or not in English. Table \ref{dataset_table} shows statistics about our dataset.\footnote{We will share the script used for dataset formulation along with links in the camera ready version of the paper} Due to the high cost of running GPT models (for zero-shot generation and G-Eval), we randomly select 50 samples for each personality in the test set.

\renewcommand{\arraystretch}{1.2}
\setlength{\tabcolsep}{4pt}
\begin{table}
\centering \scriptsize
\begin{tabular}{llll}
\hline
\textbf{Model} & \textbf{\makecell[l]{Correct \\Caption}} & \textbf{\makecell[l]{Incorrect \\Caption}} & \textbf{Difference}\\
\hline
ViT-B/32 & 0.7381 & 0.3680 & 0.3701 \\
FT ViT-B/32 (bs:1000) & 0.8329 & 0.4384 & 0.3945 \\
FT ViT-B/32 (bs:1200) & 0.8307 & 0.4076 & 0.4231 \\
FT ViT-B/32 (bs:1500) & 0.8685 & 0.4432 & 0.4253 \\
FT ViT-B/32 (bs:1600) & 0.8723 & 0.4426 & 0.4297 \\
\hline
\end{tabular}
\caption{CLIPScore (FT: Finetuned, bs: Batch size)}
\label{clipscore_table}
\end{table}

\section{Evaluation Metric}
\label{sec:bibtex}
\textbf{CLIPScore: Image and caption similarity: }To ensure the captions generated by the system are relevant to the image, we utilize CLIPScore \cite{hessel2021clipscore}, a reference-free metric that measures the quality of image captioning models and exhibits a strong correlation with human judgments.  The first row in Table \ref{clipscore_table} shows the CLIPScore (model: \textit{ViT-B/32}) of the correct image-caption vs. incorrect image-caption pairs from the test set. We observe that the score for the correct pair is relatively low, and the difference between the correct and incorrect pair scores is small. This is because the CLIP \cite{radford2021learning} model was not explicitly trained to evaluate social media captions with emojis, hashtags, etc. To address this, we finetune the CLIP model (\textit{ViT-B/32}) on our Instagram dataset, enabling it to better assess Instagram image-caption pairs. Other rows in Table \ref{clipscore_table} show the CLIPScore of finetuned CLIP model. As we increase batch size, the correct image-caption pair's score increases, and the difference between the correct and incorrect pair scores also increases. Thus, we use a finetuned \textit{ViT-B/32} model with the batch size 1600 for all our evaluations.

\begin{figure}[ht]
\begin{minipage}[b]{0.45\linewidth}
\centering
\includegraphics[width=\textwidth]{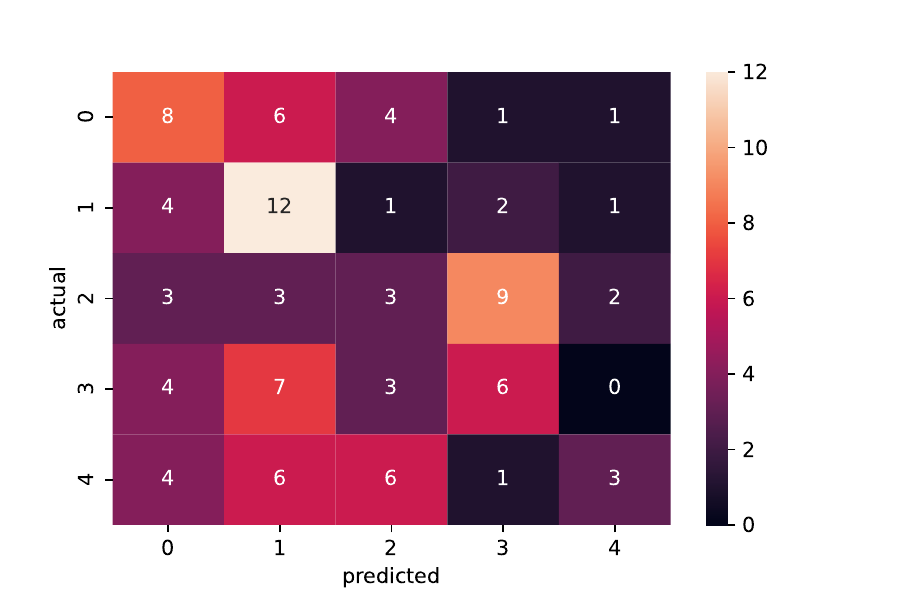}
\caption{Heatmap of human evaluations. (0: Sincerity, 1: Excitement, 2: Competence, 3: Sophistication, 4: Ruggedness}
\label{heatmap_human}
\end{minipage}
\hspace{0.5cm}
\begin{minipage}[b]{0.45\linewidth}
\centering
\includegraphics[width=\textwidth]{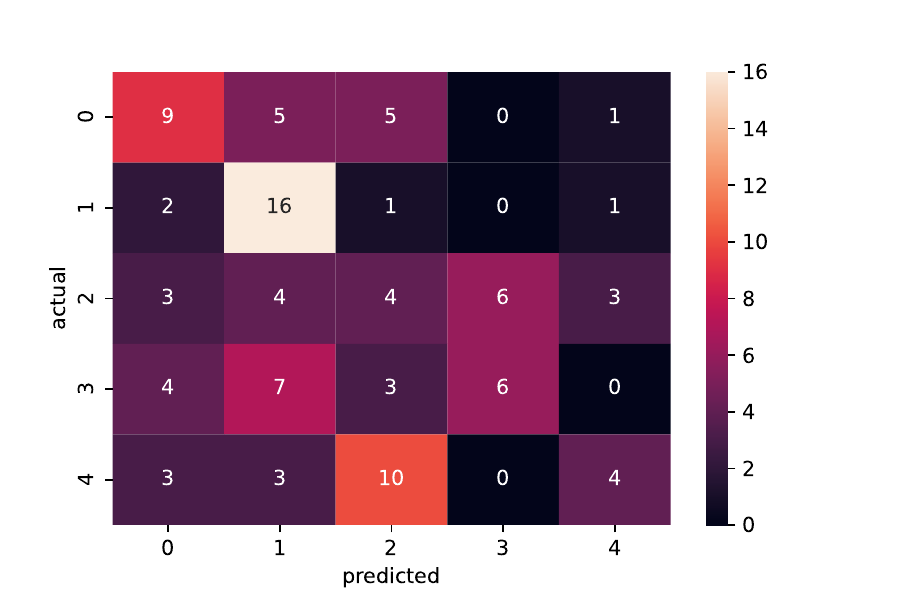}
\caption{Heatmap of G-Eval evaluations. (0: Sincerity, 1: Excitement, 2: Competence, 3: Sophistication, 4: Ruggedness}
\label{heatmap_geval}
\end{minipage}
\end{figure}

\textbf{G-Eval: Brand Personality: } Our test data is collected assuming all posts' captions belong to the brand's personality. However, we found this is not always true on manual inspection. While most of the caption reflects the brand's personality, some don't. To further investigate this, we conducted a human evaluation on a subset of test set comprising 100 captions, with 20 captions per brand. Multiple independent annotators, who were college graduates or working professionals with a strong command of English, assessed each caption based on detailed instructions. Surprisingly, we found that humans agree with our assumption only 32\% of the time. This suggests that we can't use our assumption that all the captions in each brand will reflect the brand's personality. Figure \ref{heatmap_human} shows the heatmap of human study. It compares actual annotation based on our assumption and human-predicted class. 

Since the annotations of our automatically created data are unreliable, we need a reference-free metric that agrees with human annotation. To this end, we use G-Eval \cite{liu2023geval}, a recent GPT-based metric for generation tasks that correlates highly with human judgment. G-Eval uses large language models with chain-of-thoughts(CoT) and a form-filling paradigm to assess the quality of NLG outputs. To compare the performance of G-Eval (using \textit{gpt-3.5-turbo}) against human annotation, we ran it on the same hundred data points. Due to space limitations, we do not include the prompt and hyper-parameter of G-Eval; more details are present in the appendix. Figure \ref{heatmap_geval} shows the heatmap of G-Eval, which is very similar to the human evaluation heatmap. Computing the mean squared error between the flattened heatmaps yields a low value 2.72. This suggests G-Eval with \textit{gpt-3.5-turbo} has a high correlation with human judgment, and thus, we propose using it to evaluate brand personality.

We took three disjoint subsets of training data containing 250 samples each to understand the training data better. On running G-Eval on these three subsets, we got an average accuracy of 43.33\%. Our training data is thus noisy, and a fine-tuned FlanT5 model will have a maximum accuracy of 43.33\%. If the dataset's quality improves, a fine-tuned model will be more aligned with brand personality with a higher G-Eval score.




\renewcommand{\arraystretch}{1.2}
\setlength{\tabcolsep}{4pt}
\begin{table*}[t!]
\centering\small
\begin{tabular}{llllllllll}
\hline
\textbf{Model} & \textbf{CLIPScore} & \textbf{\makecell[l]{CLIPScore \\w/o Add. \\Info}} & \textbf{\makecell[l]{G-Eval \\Accuracy}} & \textbf{\makecell[l]{G-Eval \\F1}} & \textbf{C.S.} & \textbf{\makecell[l]{hashtag}} & \textbf{entities} & \textbf{\makecell[l]{usernames}} & \textbf{URLs}\\
\hline
S. FlanT5 & 0.830 & 0.787 & 42.339 & 0.337 & 0.591 & \textbf{88.710} & 74.468 & 91.358 & 81.818 \\
N.S. FlanT5 & 0.842 & 0.810 & 38.956 & 0.367 & 0.594 & \textbf{88.710} & \textbf{76.064} & \textbf{95.679} & \textbf{100.000} \\
S. zero-GPT & 0.887 & 0.875 & \textbf{52.000} & \textbf{0.513} & 0.581 & 82.258 & 39.362 & 75.309 & 27.273\\
S. one-GPT & \textbf{0.905} & \textbf{0.898} & 45.200 & 0.439 & 0.610 & 85.484 & 62.234 & 88.272 & 54.545 \\
S. two-GPT & 0.897 & 0.890 & 38.400 & 0.370 & 0.617 & 79.032 & 62.234 & 87.037 & 54.545\\
S. three-GPT & 0.904 & \textbf{0.896} & 43.200 & 0.418 & 0.621 & 82.258 & 60.638 & 85.185 & 54.545\\
S. four-GPT & 0.893 & 0.884 & 42.800 & 0.422 & \textbf{0.627} & 80.645 & 62.234 & 83.333 & 27.273\\
N.S. zero-GPT & 0.901 & 0.877 & 49.398 & 0.403 & 0.602 & 85.484 & 47.340 & 92.593 & 45.455 \\
N.S. one-GPT & 0.892 & 0.883 & 43.373 & 0.366 & 0.590 & 80.645 & 59.043 & 87.654 & 45.455\\
N.S. two-GPT & 0.898 & 0.882 & 41.200 & 0.345 & 0.605 & 77.419 & 61.702 & 91.975 & 54.545\\
N.S. three-GPT & 0.886 & 0.873 & 42.339 & 0.356 & 0.611 & 90.323 & 60.106 & 90.741 & 45.455 \\
N.S. four-GPT & 0.886 & 0.874 & 37.600 & 0.316 & 0.617 & 83.871 & 60.106 & 88.889 & 72.727 \\
S. InstructBLIP & 0.878 & 0.877 & 39.600 & 0.346 & 0.358 & 0.000 & 0.617 & 1.064 & 0.000\\
N.S. InstructBLIP & 0.874 & 0.861 & 40.800 & 0.355 & 0.366 & 1.613 & 0.000 & 3.191 & 0.000\\
Flamingo & 0.682 & - & 42.972 & 0.354 & 0.254 & - & - & - & -\\
\hline
\end{tabular}
\caption{Quantitative Results | Add.: Additional | C.S.: Cosine Similarity | S.: Selective | N.S.: Non Selective | Columns hashtags, entities, usernames, and URLs represent the percentage of captions where these attributes were provided and had in the output.}
\label{quantitative_results}
\end{table*}

\textbf{Semantic Similarity With The Ground Truth Caption: }We evaluate the semantic similarity between the generated captions and the ground truth caption using cosine similarity of their embeddings. We use the sentence transformer library \footnote{https://github.com/UKPLab/sentence-transformers} and \textit{all-mpnet-base-v2} model \cite{reimers-2019-sentence-bert} to get embeddings of the captions. 

\textbf{Percentage of Additional Attributes in the Output: }
Our system allows users to provide additional attributes like hashtags, URLs, named entities, and usernames. We also get the percentage of captions with all the additional attributes the user provides.

\renewcommand{\arraystretch}{2}
\setlength{\tabcolsep}{6pt}
\begin{table*}[htp]
\centering \small
\begin{tabular}{|c|c|}
\hline
\raisebox{-70pt}[0pt][0pt]{\includegraphics[width=60mm]{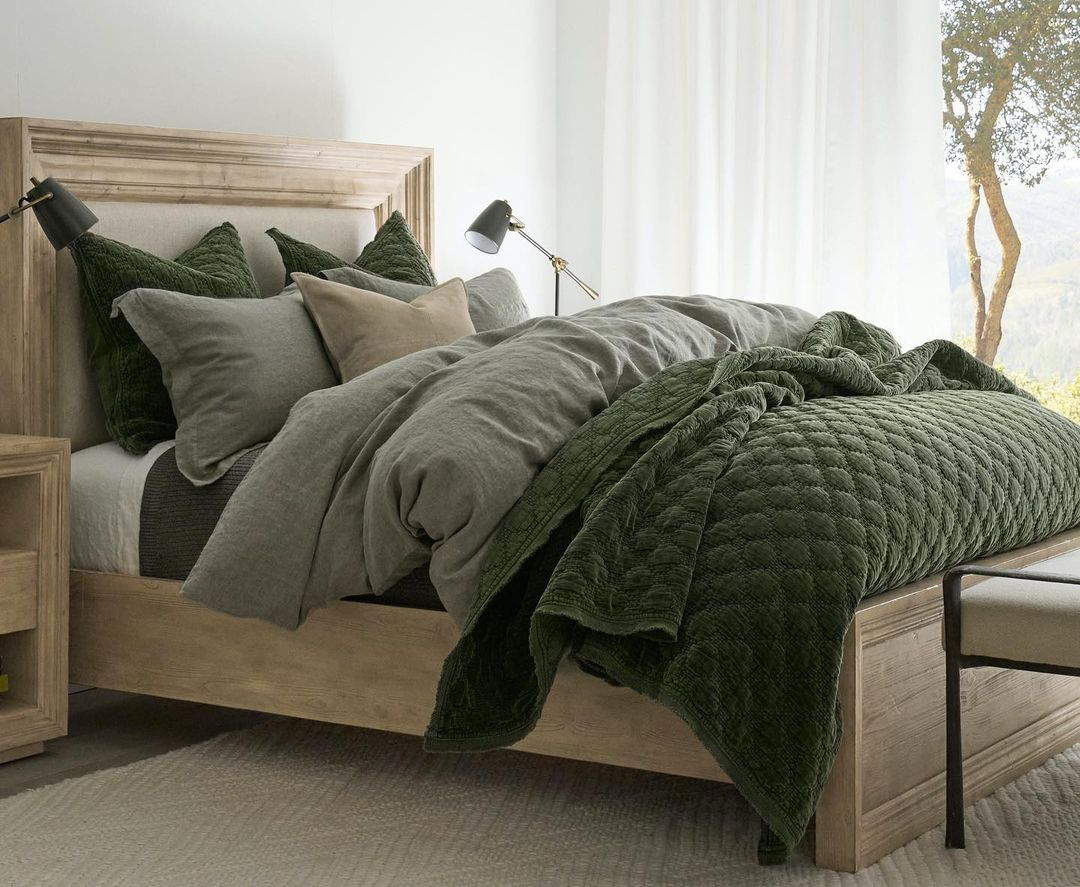}} & 
\makecell[l]{
\textbf{Our (FlanT5):}  What's your favorite way to cozy up \\ this season? We're bedding in on the emerald hue.\\
\textbf{Our (GPT):}  Dreaming of cozy nights in this season\\
\textbf{Flamingo: } Apparently it's no joke they can \\
\textbf{InstructBLIP: }, the bedroom is adorned with a \\ green bedspread and matching curtains \\
\textbf{Actual Caption: }A bedding palette for every mood \\ Swipe for a round-up of the bedroom looks we \\ love this season! Which palette are you? Tell us in the \\comments!\\
\textbf{BLIP2 Caption: }a bed with a green comforter and \\ a wooden headboard \\
\textbf{Target personality: }sophistication \\
\textbf{Additional Attributes: } \\Named Entities: 'this season'
}
\\
\hline

\raisebox{-40pt}[0pt][0pt]{\includegraphics[width=65mm]{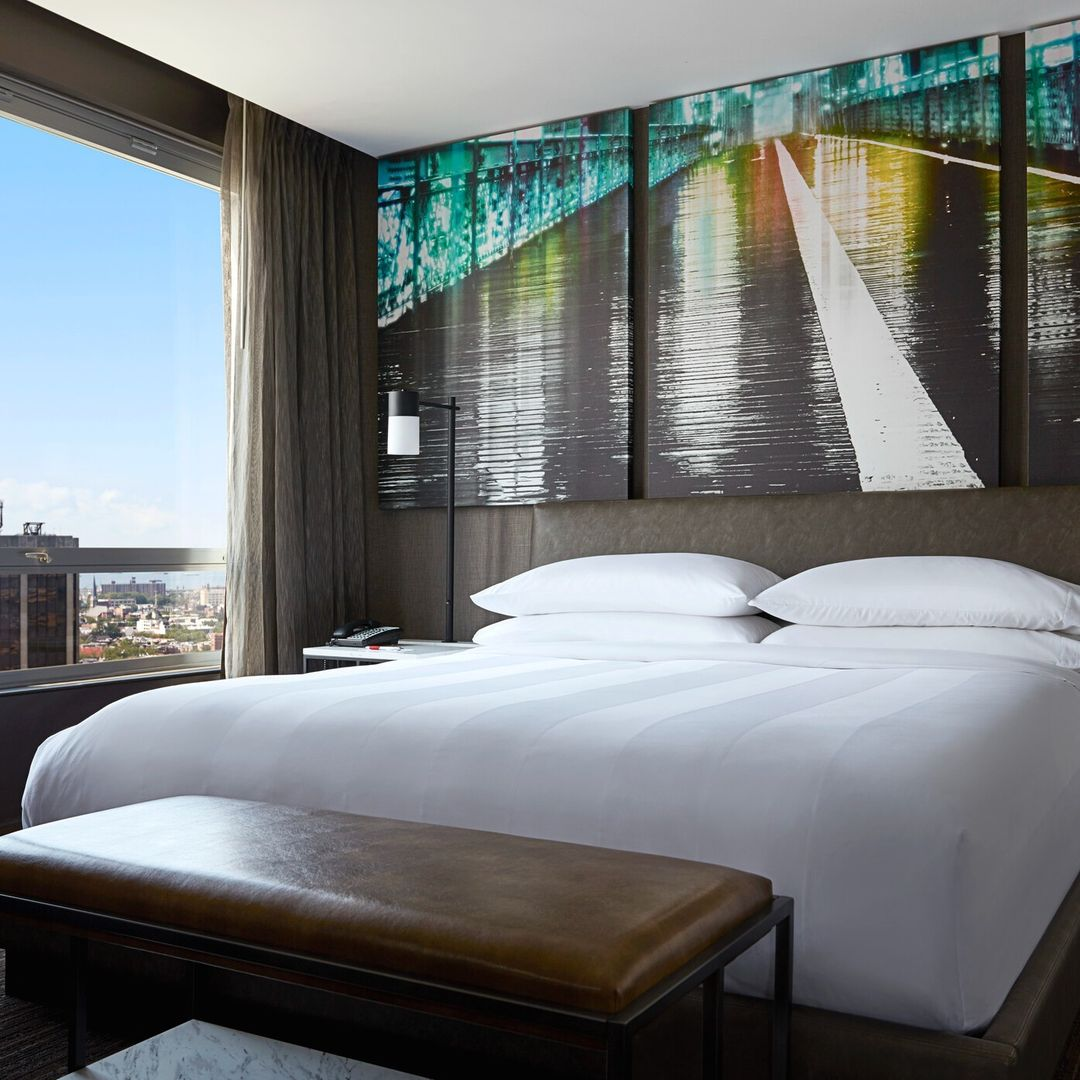}} & 
\makecell[l]{
\textbf{Our (FlanT5):}  Staying in New York City in the fall? \\This fall @brooklynmarriott has a view for you. \\Looks like a great place to crash. Check in for a \\ chance to win a staycation with a view of the \\Brooklyn Bridge!\\
\textbf{Our (GPT):}  Waking up to this stunning view of \\New York City in the fall is truly unforgettable. \\Our hotel room at @brooklynmarriott offers the \\ perfect vantage point to admire the beauty of the city, \\with the iconic Brooklyn Bridge in the distance. Can \\you imagine a better way to start your day? \\
\textbf{Flamingo: } Forty-eight hours. In the week leading \\ up to Sandalwood and Chartreuse's wintertime \\dedication, this is the amount of time that \\inside audible hoots (have you heard Chartreuse \\yodeling?) replaced the sound of carpenters, \\furniture makers, wood roughing \\
\textbf{InstructBLIP: }a bedroom with a large bed and a \\painting of a cityscape on the wall \\
\textbf{Actual Caption: }There is nothing quite like New \\York City in the fall. The weather is perfect for a \\short walk to the Brooklyn Bridge when you stay \\ at @brooklynmarriott.\\
\textbf{BLIP2 Caption: }a large bed in a hotel room with \\a view of the city \\
\textbf{Target personality: }competence \\
\textbf{Additional Attributes: } \\Named Entities: 'New York City', 'the fall', \\'the Brooklyn Bridge' \\
Usernames: '@brooklynmarriott'
}
\\
\hline
\end{tabular}
\caption{\label{qualitative_example}
Qualitative Examples
}
\end{table*}

\section{Result and Discussion}
\textbf{Implementation Details: }We use the pretrain\_flant5xxl BLIP2 model for automatic image captioning. We finetune FlanT5-XL for Instagram caption generation. The finetuning is done for 80k steps, and we choose the model with the best performance on the validation set. The batch size is $4$, with a learning rate of $0.00001$. The training is done on one $A100$ $80$GB GPU and takes around $50$ hours for the entire training. We use gpt-$3.5$-turbo with temperature $0.7$, top\_p $0.95$, frequency\_penalty $0$, and presence\_penalty $0$ for image captioning and G-Eval.

\textbf{Baseline}
We compare our results with two recent state-of-the-art few-shot visual language models: Flamingo \cite{alayrac2022flamingo} and InstructBLIP \cite{instructblip}. We use Flamingo-9B with two-shot inference and InstructBLIP Vicuna-13B in a one-shot setting. The input to these models is the image and a prompt for generating social media captions. We try both selective and non-selective variants of prompt with InstructBLIP. We experiment with various prompt variants, with no significant difference in results. So, we used the same prompt as the one-shot GPT call (Table \ref{gpt_example}). We do not provide additional attributes to Flamingo-9B; we use it as a baseline to see how good it is in Instagram Caption generation aligned with brand personality.

\textbf{Quantitative Results: }Table \ref{quantitative_results} presents the quantitative results. The Flamingo-9B model performs poorly in terms of CLIPScore and cosine similarity, indicating that its generated captions are not grounded in the original image and ground truth caption. In fact, captions generated by Flamingo-9B are gibberish. Captions generated by both variants of InstructBLIP have high CLIPScore and G-Eval scores. However, almost no additional attributes the user provides are present in the output. As shown in the next section, all the captions generated by InstrucBLIP are plain English sentences despite trying different paraphrases of the prompts. Such English sentences are not useful for the social media campaign. Thus, the current vision-language models are unsuitable for our task. Additionally, these models require users always to have an image; however, our framework allows users to generate captions even if they don't have an image. They can provide a one-line image description and use the second part of our framework to generate the final social media caption. 

Selective variants performed better than the non-selective variants for both Finetuned FlanT5 and GPT-based models. The selective model does not feed any extra information to the models, and the instructions are easy to understand, and thus it performs better. As we increase the number of examples in the GPT model, the generated captions are more similar to the ground truth caption, but the G-Eval accuracy decreases. Also, as we increase the number of examples, the percentage of additional attributes in the output increases. Thus, based on quantitative results, we recommend using zero-shot or one-shot selective variant of GPT if the training data is low.

We observe that the selective variant of the finetuned FlanT5 model generates captions that are more grounded to the image, grounded to the true caption, have more G-Eval score, and cover more attributes in the output than non-selective variant. Thus if privacy or cost is a concern and we have sufficient training data, we recommend using a selective variant of finetuned FlanT5 model, otherwise we recommend using one-shot selective variant of GPT. In section 5.2, we discussed that the average G-Eval accuracy of the training dataset is 43.33\%. Despite that, our finetuned FlanT5 model achieves G-Eval accuracy of 42.34\%; this further suggests that if we have a more clean dataset, the G-Eval accuracy will increase without affecting other metrics. In this work, we are limited by the noise in the dataset; however, for future work with a clean dataset, we recommend finetuning a FlanT5 model over zero/few-shot GPT calls.

\textbf{Qualitative Results: }
Table \ref{qualitative_example} shows some qualitative results. We show the output of selective variants for all the models. The output by InstructBLIP is just a single-line description which is more suited for image captioning than Instagram captioning. The output of the Flamingo model is hallucinated, which explains its low cosine similarity and CLIPScore. The overall output of both FlanT5 and GPT3 models are grounded to the image, in the target personality, and contains additional attributes the user provides.

\section{Conclusion}
In this work, we propose a new task of \textbf{generating automatic brand captions from the brand images for social media posts} while aligning with the \textbf{brand-specific personalities}. We quantitatively study the limitation with existing literature, dataset, and evaluation metrics. We address these limitations in our work. Our framework allows users to provide an image or description of an image along with the target personality and brand-specific attributes. We compare Finetuning open-source model and GPT outputs and provide insights into the working of each. This work will enable future researchers in marketing and multimodal Instagram caption generation.

\section*{Limitations}
The main limitation of our work is the availability of good-quality datasets. While we try to understand and address these limitations, future work should focus on creating richer and high-quality datasets. Such a quality dataset will help in evaluations and finetuning the FlanT5 model with high accuracy. We extensively use GPT for evaluations, which is costly, thereby limiting us to a small test dataset.
\bibliography{anthology,custom}

\clearpage

\appendix
\section*{Appendix}
\section{Brands Present in Each Personality}
Table \ref{brands_table} enumerates all brands present in each personality.

\section{Instruction for FlanT5 finetuning}
Table \ref{flant5_example} shows the instruction for FlanT5 finetuning with both selective and non-selective variants.

\section{Prompt for one-shot GPT call.}
Table \ref{gpt_example} shows the prompt for a one-shot GPT call for selective and non-selective variants.

\section{G-Eval}
Table \ref{geval_example} shows the prompt to the GPT for G-Eval. We used \textit{gpt-3.5-turbo} with temperature $0.7$, top\_p $0.95$, frequency\_penalty $0$, presence\_penalty $0$ and generated 10 outputs. We selected the personality that occurred maximum times.

\renewcommand{\arraystretch}{2}
\setlength{\tabcolsep}{6pt}
\begin{table}
\centering \scriptsize
\begin{tabular}{ll}
\hline
\textbf{Personality} & \textbf{Brands}\\
\hline
Sincerity & \makecell[l]{Cracker Barrel, IHOP, Pet Smart, \\Campbell Soup, Whole foods, and \\Walgreens (\textbf{Test})} \\
\hline
Excitement & \makecell[l]{Urban Outfitters, Forever 21, EA, \\Progressive Insurance, The children's \\place, and Mattel (\textbf{Test})} \\
\hline
Competence & \makecell[l]{UPS, Microsoft, GE, Autozone, Sony, \\and Marriott (\textbf{Test})} \\
\hline
Sophistication & \makecell[l]{Bloomingdale's, Revlon, Audi, Ann \\Taylor, L'Oreal, and Pottery Barn (\textbf{Test})}\\
\hline
Ruggedness & \makecell[l]{Columbia Sportswear, The North Face, \\General Motor's, Advance Auto Parts, \\Lowe's, and Goodyear (\textbf{Test})} \\
\hline
\end{tabular}
\caption{Brands in each personality}
\label{brands_table}
\end{table}

\renewcommand{\arraystretch}{2}
\setlength{\tabcolsep}{6pt}
\begin{table*}
\centering\small
\begin{tabular}{|ll|l|}
\hline
\textbf{Image Caption} & \makecell[l]{a woman wearing a face mask \\speaks to a crowd} & \multirow{8}{*}{\makecell[l]{\textbf{Selective} \\
Create an Instagram caption from the following text. The tone of \\the text should be sincere. Make use of named entities, usernames \\ at the end.\\
a woman wearing a face mask speaks to a crowd\\
Named Entities: Kamala Harris, last week, Roz Brewer, today. \\Usernames: @vp, @wba\_global.
\\ \\
\textbf{Non-selective} \\
Create an Instagram caption from the following text. The tone of \\the text should be sincere. Make use of named entities, links, \\hashtags and usernames present at the end.\\
a woman wearing a face mask speaks to a crowd\\
Named Entities: Kamala Harris, last week, Roz Brewer, today. \\Links: None. Hashtags: None. Usernames: @vp, @wba\_global.}} \\
\textbf{Named Entities} & \makecell[l]{Kamala Harris, last week, \\Roz Brewer, today} & \\
\textbf{Usernames} & @vp, @wba\_global & \\
\textbf{Hashtag} & - & \\
\textbf{Links} & - & \\
& & \\
& & \\
& & \\ \hline

\end{tabular}
\caption{\label{flant5_example}
Instruction for FlanT5 finetuning. We experiment with two variants, selective and non-selective. In selective, if some attribute is not provided, it is not mentioned in the instruction. In contrast, in non-selective, if some attribute is not provided, it is mentioned in the instruction with a None value.
}
\end{table*}

\renewcommand{\arraystretch}{2}
\setlength{\tabcolsep}{6pt}
\begin{table*}
\centering\small
\begin{tabular}{|l|}
\hline
\makecell[l]{\textbf{Selective}\\Create an Instagram caption from the provided text. The tone of the text should be sophisticated. If you are \\provided with addtional information, make use of them. These additional information could be named entities, \\links, hashtags and usernames. Do not create named entities, links, hashtags or usernames by yourself. \\Text: a pair of metallic sneakers on a white couch\\Usernames: @balenciaga. \\Instagram caption: No shoes on the couch (unless they're @balenciaga) \\ \\ Text: a dog sleeping on a couch in a living room \\Named Entities: PB Comfort Sofa. \\Hashtags: \#dogsofinstagram, \#doglovers, \#petapproved, \#livingroominspo, \#homedetails, \#homegoals, \\ \#homeinspo, \\ \#potterybarn. \\Usernames: @apriljoy\_ful. \\Instagram caption:\\ \\ \\ \textbf{Non Selective} \\ Create an Instagram caption from the provided text. The tone of the text should be sophisticated. Only make use \\ of named entities, links, hashtags and usernames present at the end and nothing else. If you see None in front \\of named entities, links,  hashtags or usernames, do not include anything in the output. \\ Text: a pair of metallic sneakers on a white couch \\ Named Entities: None. Links: None. Hashtags: None. Usernames: @balenciaga. \\ Instagram caption: No shoes on the couch (unless they're @balenciaga) \\ \\ Text: a dog sleeping on a couch in a living room \\ Named Entities: PB Comfort Sofa. Links: None. Hashtags: \#dogsofinstagram, \#doglovers, \#petapproved, \\\#livingroominspo,  \#homedetails, \#homegoals, \#homeinspo, \#potterybarn. Usernames: @apriljoy\_ful. \\ Instagram caption: \\ \hline
}
\end{tabular}
\caption{\label{gpt_example}
Prompt for one-shot GPT call. We experiment with two variants, selective and non-selective. In selective, if some attribute is not provided, it is not mentioned in the prompt. In contrast, in non-selective, if some attribute is not provided, it is mentioned in the prompt with a None value.
}
\end{table*}

\renewcommand{\arraystretch}{2}
\setlength{\tabcolsep}{6pt}
\begin{table*}
\centering\small
\begin{tabular}{|l|}
\hline
\makecell[l]{You will be given an instagram caption. Your task is to predict the brand personality from the caption. The possible brand \\personalities are: sincerity, excitement, competence, sophistication, and ruggedness.\\You will be provided with the definition of each personality type and evaluation steps. Please make sure you read and \\understand the evaluation steps carefully. Please keep this document open while reviewing, and refer to it as needed.\\ \\ Sincerity: The attributes represented by this brand personality are down-to-earth, family-oriented, small-town, honest, \\sincere, real, wholesome, original, cheerful, sentimental, and friendly. \\ Excitement: The attributes represented by this brand personality are trendy, exciting, spirited, cool, young, imaginative,\\ unique, up-to-date, independent, and contemporary.\\Competence: The attributes represented by this brand personality are reliable, hard-working, secure, intelligent, \\technical, corporate, successful, leader, and confident.\\Sophistication: The attributes represented by this brand personality are upper-class, glamorous, good-looking, charming, \\feminine, and smooth.\\Ruggedness: The attributes represented by this brand personality are masculine, outdoorsy, tough, rugged, and western.\\ \\ Evaluation Steps: \\1. Read and understand the Instagram caption: Carefully read the provided caption and try to grasp the overall tone and \\theme of the message.\\2. Identify keywords and phrases: Look for specific words or phrases in the caption that can be related to the attributes \\of the brand personalities.\\3. Create a list of possible brand personalities: Based on the keywords and phrases identified, create a list of possible \\brand personalities that the caption could represent.\\4. Analyze the caption for sincerity: Check if the caption demonstrates attributes such as down-to-earth, family-oriented, \\small-town, honest, sincere, real, wholesome, original, cheerful, sentimental or friendly. If it does, the brand personality \\could be sincerity.\\5. Analyze the caption for excitement: Check if the caption demonstrates attributes such as trendy, exciting, spirited, \\cool, young, imaginative, unique, up-to-date, independent or contemporary. If it does, the brand personality could \\be excitement.\\6. Analyze the caption for competence: Check if the caption demonstrates attributes such as reliable, hard-working, \\secure, intelligent, technical, corporate, successful, leader or confident. If it does, the brand personality could be \\competence.\\7. Analyze the caption for sophistication: Check if the caption demonstrates attributes such as upper-class, glamorous, \\good-looking, charming, feminine or smooth. If it does, the brand personality could be sophistication.\\8. Analyze the caption for ruggedness: Check if the caption demonstrates attributes such as masculine, outdoorsy, tough, \\rugged or western. If it does, the brand personality could be ruggedness.\\9. Do note that the caption might not have all the attributes.\\10. Evaluate the overall impression: Consider the overall impression of the caption, including the emotional tone and \\visual elements (such as emojis, hashtags).\\11. Make a final decision: Based on the analysis and overall impression, determine the most likely brand personality that \\the caption represents.\\ \\ Example: \\Instagram caption: \#\#\_\_caption\_\_\#\#\\Brand personality (select from Sincerity, Excitement, Competence, Sophistication, and Ruggedness):} \\
\hline
\end{tabular}
\caption{\label{geval_example}
Prompt for G-Eval. We replace \textit{\#\#\_\_caption\_\_\#\#} with the caption to be evaluated.}
\end{table*}

\end{document}